\documentclass[conference]{IEEEtran}
\IEEEoverridecommandlockouts

\usepackage{cite}
\usepackage{amsmath,amssymb,amsfonts}
\usepackage{algorithmic}
\usepackage{graphicx}
\usepackage{textcomp}
\usepackage{xcolor}
\usepackage{tikz} 
\usepackage{pgfplots}
\usepackage{csquotes}
\usepackage{subcaption}
\usepackage{authblk}

\def\BibTeX{{\rm B\kern-.05em{\sc i\kern-.025em b}\kern-.08em
    T\kern-.1667em\lower.7ex\hbox{E}\kern-.125emX}}
\usepackage{hyperref}
\newcommand{\smallheading}[1]{\vspace{0.1cm}\noindent\textbf{#1}\ \ }
\begin{document}

\title{PIER: A Novel Metric for Evaluating What Matters in Code-Switching}

\author{\textit{Enes Yavuz Ugan}\textsuperscript{$\star$}}
\author{\textit{Ngoc-Quan Pham}\textsuperscript{$\star\dagger$}}
\author{\textit{Leonard Bärmann}\textsuperscript{*}}
\author{\textit{Alex Waibel}\textsuperscript{†}}
\vspace{30pt}
\setlength{\affilsep}{0.3em}
\affil{\textsuperscript{*}Interactive Systems Lab, Karlsruhe Institut of Technology (KIT), Germany}
\affil{\textsuperscript{†}InterACT, Carnegie Mellon University (CMU), USA}

\affil{\texttt{enes.ugan@kit.edu}}


\maketitle

\begin{abstract}
Code-switching, the alternation of languages within a single discourse, presents a significant challenge for Automatic Speech Recognition. 
Despite the unique nature of the task, performance is commonly measured with established metrics such as Word-Error-Rate (WER).
However, in this paper, we question whether these general metrics accurately assess performance on code-switching. 
Specifically, using both Connectionist-Temporal-Classification and Encoder-Decoder models, we show fine-tuning on non-code-switched data from both matrix and embedded language improves classical metrics on code-switching test sets, although actual code-switched words worsen (as expected).
Therefore, we propose Point-of-Interest Error Rate (PIER), a variant of WER that focuses only on specific words of interest.
We instantiate PIER on code-switched utterances and show that this more accurately describes the code-switching performance, showing huge room for improvement in future work.
This focused evaluation allows for a more precise assessment of model performance, particularly in challenging aspects such as inter-word and intra-word code-switching.

\end{abstract}


\section{Introduction}
\label{sec:intro}
Code-switching (CSW), a pervasive linguistic phenomenon, can manifest in a variety of forms\footnote{By definition, CSW, sometimes also called code-mixing, can be viewed differently in different areas of linguistic studies \cite{ezeh2022code}.}.
Typically, it involves the use of a foreign \enquote{embedded} language within the main \enquote{matrix} language of the discourse, such as mixing in English words or phrases into a German conversation.
Naturally, this poses a formidable challenge for all natural language processing (NLP) \cite{zhang2023multilingual} tasks, with Automatic Speech Recognition (ASR) \cite{yan2023towards,biswas2022code} being one outstanding example.
Nevertheless, mastering CSW is of greatest importance, as usage of a second language can have a lot of socio-linguistic importance and convey meaningful information \cite{scotton1977bilingual, dougruoz2023survey} that should not be neglected for downstream NLP applications \cite{ugan2023modular, Hourin_Binder_Yaeger_Gamerdinger_Wilson_Torres-Smith_2013, waibel2012simultaneous, waibelsltu201045, huber2023end, waibe112005chil}.

Multilingual speech recognition research has evolved significantly over time.
Beginning with classical models~\cite{stuker2003multilingual, muller2017language}, to nowadays prevalent End-to-End trained neural networks \cite{abdallah2024leveraging}.
Recent works on CSW cover language pairs such as Mandarin-English \cite{aditya2024attention, liu2024enhancing}, Arabic/Tunisian-English \cite{abdallah2024leveraging,kadaoui-etal-2024-polywer,hamed2023benchmarking} and German-English \cite{ugan2022language,huber2022code}.


While prior research has acknowledged the complexity of code-switching concerning data collection \cite{dogruoz-etal-2023-representativeness, ugan2024decm} and processing \cite{cetinoglu-etal-2016-challenges}, in this paper, we ask:
\emph{How to meaningfully isolate the problem of code-switching to accurately measure its performance?}
Although Word- and Character-Error-Rate (WER/CER) are established and meaningful metrics for ASR evaluation, we observe that they are in fact not sufficient to evaluate the true code-switching performance, and can even be misleading.
These common metrics underweigh the actually important points of code-switching speech by mixing them with the performance on the (dominant) matrix language.
Our experiments prove that the prevalence of this problem is independent of language or model architecture.
Specifically, we fine-tune different competitive models on various language pairs using mixed data from matrix and embedded language.
While this training is multilingual, it does not include code-switched samples, i.e. each sample is monolingual in itself.
Intuitively, such training should not help a lot when evaluating on a CSW test set, as such samples are still out-of-distribution for the fine-tuned model.
Despite that, results using metrics such as WER show performance improvements, while the recognition of actual code-switched instances is deteriorating.

To address this shortcoming, we propose a new metric called \textbf{Point of Interest Error Rate (PIER)}, which calculates Error Rate only considering words at interesting positions.
We instantiate this with treating the embedded language's words as points of interest, and show that PIER actually increases when applying the fine-tuning described above, accurately describing the worsened CSW performance.
PIER is easily applicable when the points of interest can be identified, which is the case for most available CSW test sets, as they either consist of different writing scripts or provide additional language annotations.
Language pairs such as Arabic/German \& English can lead to morphological mixing of those languages within a word, which we refer to as intra-word CSW.
PIER also enables the analysis of such intra-word CSW by defining only those words as points of interest.
In our experiments, we show that this task is even more challenging for current models than inter-word CSW, i.e. the mixing of monolingual words within a utterance.

Similarly, researchers across various domains have highlighted the limitations of existing evaluation metrics and proposed new ones to improve model assessment. Examples include tasks like lip synchronization \cite{yaman2024audio}, and supervised quality estimation for machine translation (MT) \cite{dinh2023perturbation}. Papers like \enquote{A call for clarity} \cite{post2018call, schmidt2022non} have contributed to standardizing MT evaluation. Similarly, \cite{gaido2024speech} points out that BLEU, a standard MT metric, disadvantages large language models, suggesting the need for additional metrics.

To summarize, by evaluating with our proposed PIER metric, researchers working on CSW can better understand whether a model's performance gains result from better monolingual capabilities of the model or truly improved CSW ability. To the best of our knowledge, this is the first attempt to separate code-switching specific errors within a test set enabling us to identify and quantify the errors for intra-word and inter-word types of CSW in several datasets.
Find our open-source implementation here\footnote{\url{https://github.com/enesyugan/PIER-CodeSwitching-Evaluation}}.

\section{Experimental Setup}
\label{sec:setup}
We employ recent competitive models whisper-large-v3 (W-large) and whisper-small (W-small) \cite{radford2023robust} to show our results are independent of the initial model's performance.
Additionally, we employ the massively multilingual MMS model \cite{pratap2024scaling}, a connectionist-temporal-classification (CTC) model.

\label{subsec:data}
\smallheading{Data} Table~\ref{tab:test-data} shows the publicly available and widely used datasets that we use, ensuring the diversity and representativeness of our test cases.
They cover a wide range of linguistic scenarios for code-switching:
In Fisher, data has the same writing script but differing language families.
Arzen consists of code-switching between two languages of different families, also utilizing different writing scripts.
Datasets for Mandarin-English code-switching such as SEAME showcase different language families and not just different writing scripts but also very different script semantics.
By using such diverse data we highlight that our findings are not just language- or dataset-specific but a general phenomenon over all code-switching datasets and languages.
\begin{table}[t!]
\caption{\label{tab:test-data} Test data.}
\small
\centering
\begin{tabular}{|ll|r|r|} \hline
 Dataset & Language &  Speech [h] & Utterances \\ 
 \hline
 \hline
    TED \cite{rousseau2012ted} & EN  & 2.62 & 1155  \\
    \hline
    MLS \cite{pratap2020mls} & DE & 14.29 & 3394  \\
    \hline
    MGB \cite{ali2016mgb} & AR & 9.58 & 5365 \\
    \hline  
    Aishell \cite{bu2017aishell} & ZH & 10.03 & 7176 \\
    \hline
    CommonVoice \cite{ardila2019common} & ES & 26.85 & 15753  \\
    \hline    
    \hline
    Fisher \cite{weller2022end} & ES-EN  & 1.63 & 986 \\
    \hline
    Arzen \cite{hamed2020arzen} & AR-EN  & 2.87 & 1470\\
    \hline
    SEAME.sge \cite{lyu2010seame} & ZH-EN  & 3.93 & 5321 \\
    \hline
    SEAME.man \cite{lyu2010seame} & ZH-EN & 7.49 & 6531 \\
    \hline
\end{tabular}

\vspace{-3mm}
\end{table}

\section{Evaluating Code-Switching with WER/CER}
In this section we want to show that only using WER or CER is not enough for evaluating code-switching performance properly. 
In case of intra-sentential code-switching there are only few words or even only one word in the embedded language. 
Therefore, improvements in terms of the matrix language can overshadow the impact of the embedded language.
For example, the surrounding words can be recognized correctly leading to a low error rate overall.
Thus, even wrong conclusions might be drawn, or at least the model's true capability of switching/predicting a foreign language inside the matrix language is not obvious.

\smallheading{Monolingual Base Scores}
\label{subsec:mono_vs_csw}
In Table~\ref{tab:eval-mono} we initially show the results of the base models used in our experiments on monolingual data.
The goal is to show that our decoding parameters and settings are well suited and yield SotA results on widely used monolingual test sets.
All our following experiments utilize the same decoding parameters.

\begin{table}[t!]
    \caption{Monolingual evaluation of pre-trained models. For Mandarin, the evaluation metric is CER, all other data sets are evaluated using WER. }
    \small
    \setlength\tabcolsep{2px}
    \renewcommand{\arraystretch}{1}
    \centering
    \begin{tabular}{|c|c|c|c|c|c|} \hline
       Model  & English & German & Arabic & Spanish & Chinese  \\
    \cline{2-6}
    & TED & MLS & MGB & CV & AISHELL \\
    \hline
    \hline
    W-large & 6.07 & 5.88 & 22.61 & 4.91 &  10.02 \\
    W-small & 6.38 & 10.58 & 35.07 & 10.24 & 24.99 \\
    MMS & 11.16 & 8.68 & 40.41 & 9.93 & 31.23 \\
    \hline
    \end{tabular}
    \label{tab:eval-mono}

\vspace{-3mm}
\end{table}

\smallheading{CSW Base Scores}
Baseline results of our pre-trained models on code-switching data are reported on the left side of Table~\ref{tab:eval-comparison-wer}.
We can see an expected performance drop compared to monolingual tests even for our competitive models trained on thousands of hours of paired data.
For more detailed analysis, we report results using different decoding strategies.
For Whisper, these include an 'agnostic' setting where the model automatically determines the language being decoded. 
The 'EN' setting specifies that the task is to transcribe English speech, while 'X' corresponds to the matrix language of each dataset (e.g., for the Fisher test set, 'X' is set to Spanish).
For the MMS model, we select either the English (EN) or the respective matrix language adapters (X) based on the dataset's matrix language.
\begin{table*}[t!]
\caption{Comparison of pre-trained and fine-tuned model performances. Fisher (ES-EN), Arzen (AR-EN) values are reported in WER/CER. MER for SEAME. Decoding X depicts the respective matrix language of the dataset. For right hand part of the table, MMS X represents fine-tuned adapter layers for CTC.}
\centering
\scriptsize
\setlength\tabcolsep{5px}
\renewcommand{\arraystretch}{1}
\begin{tabular}{|l|c|c|c|c|c||c|c|c|c|c|}
\hline
\multicolumn{1}{|c|}{\textbf{Model}} & \multicolumn{1}{c|}{\textbf{Decoding}} & \multicolumn{4}{c||}{\textbf{Pre-trained Model}} & \multicolumn{4}{c|}{\textbf{Fine-tuned Model}} \\
\hline
 &  & \textbf{Fisher} & \textbf{Arzen} & \textbf{SEAME.man} & \textbf{SEAME.sge} & \textbf{Fisher} & \textbf{Arzen} & \textbf{SEAME.man} & \textbf{SEAME.sge} \\
\cline{3-6} \cline{7-10}
 &  & WER / CER & WER / CER & MER & MER & WER / CER & WER / CER & MER & MER \\
\hline
W-large & agnostic & 29.43 / 19.17 & 52.80 / 28.78   & 31.01 & 53.81 & 27.94/ 15.96 & 39.68 / 20.41  & 36.13 & 52.90 \\
              & EN  & 42.58 / 28.59 & 103.81 / 84.38 & 59.81 & 47.79 & 27.34 / 15.21 & 102.81 / 81.82 & 35.72 & 52.03 \\
              & X  & 30.57 / 20.65 & 53.90 / 29.38  & 30.26 & 65.67 & 28.01 / 15.99 & 40.07 / 20.79 & 36.13 &  52.92 \\
\hline
W-small & agnostic & 35.98 / 22.74 & 69.16 / 41.06 & 41.17 & 68.24 & 36.77 / 21.50 & 55.08 / 31.31 & 42.09 & 87.67\\
              & EN  & 68.04 / 47.14 & 105.80 / 85.35  & 84.76 &   65.26 & 36.33 / 21.32 & 114.25 / 90.56  & 62.13  & 55.57\\
              & X  & 36.13 / 23.35 & 69.56 / 41.12  & 38.22 & 63.17 & 36.81 / 21.50 & 55.36 / 31.56 & 43.55 & 94.03\\
\hline
MMS     & EN & 50.58 / 25.13 & 96.95 / 81.13 & 95.86 & 89.20 & --& --& --& --\\
        & X & 47.29 / 24.47 & 86.44 / 49.85 & 68.25 & 87.41 & 38.87 / 18.25 & 66.48 / 34.06 & 61.19 & 91.55\\
\hline
\end{tabular}
\label{tab:eval-comparison-wer}
\vspace{-3pt}
\end{table*}
\subsection{Why WER is not enough}
In this section we want to improve our models performance on our code-switching test cases.
Fine-tuning or solely using training data, provided with the test set, is a common practice in many works in the literature~\cite{ye2024sc, yang2024effective}.
Therefore, we use the training splits that come along with the Fisher, Arzen and SEAME data sets, assuming that the training and validation/test sets are drawn from the same condition, for example recording conditions, noise levels and styles.
However, in order to better make our case, we restrict ourselves to only using monolingual utterances out of those training sets. 
This will potentially show that code-switching examples are not really needed for improving code-switching ASR.
Additionally, for each experiment, we randomly sample the same amount of monolingual English utterances from publicly available datasets such as TED or CV.
For fine-tuning, we apply early-stopping if the model does not improve after five consecutive evaluations.
WER and CER results are depicted on the right side of Table~\ref{tab:eval-comparison-wer}.
Comparing with the values of the base models on the left hand side of the table, its quite surprising that the common error rates are improved over the test sets. 
There are a few cases where the WER remains unchanged or deteriorates slightly.
The language agnostic Whisper large model (W-large) improves absolute by 1.49\% WER on the Spanish-English Fisher data.
On the Arabic-English code-switching test data (Arzen) all our models are marginally improved.
The agnostic Whisper small model improves by 14.08\% WER.

However, it is questionable if most of the improvement above comes from recognizing the main languages more accurately while the minority of code-switched words are ignored.
To answer this question, we suggest to evaluate models' code-switching performance by looking at the Point of Interest Error Rate (PIER).

\section{Point of Interest Error Rate}
\label{subsec:PIER}
Let $\mathcal{I}$ be the set of indexes (positions) of points of interest, i.e. $\mathcal{I} = \{i_1, i_2, \dots, i_n\}$, $n\leq {|REF|}$.  
We compute an alignment $\mathcal{A}$ 
between two strings, \(\text{REF}\) and \(\text{HYP}\), which produces a set of edit operations. Each edit operation is characterized by:
\[
o = (\;t\;, i_\text{src}\;, i_\text{res}\;)
\]
with,
\(t \in \{\text{substitution}, \text{insertion}, \text{deletion}\}\), \(i_\text{src}\) being the position of the word in \(\text{REF}\), \(i_\text{res}\) being the position of the word in \(\text{HYP}\).

\noindent We define the subset \(\mathcal{A}_{\mathcal{I}}^{*}\) of the alignments  \(\mathcal{A}\) as:
\[
\mathcal{A}_{\mathcal{I}}^{*} = \left\{ o \in \mathcal{A} \mid i_\text{src}^o \in \mathcal{I} \right\}
\]
Additionally, if the last index in \(\mathcal{I}\) corresponds to the last word in \(\text{REF}\), we also include all alignments where \(i_\text{src}\) is greater than the last index in \(\mathcal{I}\):
\[
\mathcal{A}_{\mathcal{I}} = 
\begin{cases} 
\mathcal{A}_{\mathcal{I}}^{*} \cup \left\{ o \in \mathcal{A} \mid i_\text{src}^o > |\text{REF}| \right\}, &\max(\mathcal{I}) = |\text{REF}| \\
\mathcal{A}_{\mathcal{I}}^{*}, & \text{otherwise}
\end{cases}
\]
Finally, we count the operations in \(\mathcal{A}_{\mathcal{I}}\) and normalize by the total number of indexes (words) we are interested in.
\[
\text{PIER}= \frac{|\mathcal{A}_{\mathcal{I}}|}{|\mathcal{I}|}
\]

The idea of PIER is to stay as close to the established WER as possible, to have a similarity to known and proven metrics while also focusing on the actual capabilities of code-switching.
This way, during evaluation, we do not only get the amount of miss-transcribed words of the embedded language, but we can also get which errors were associated with points of code-switching.
As an example, while one model might just substitute English words with literal transliterations of Mandarin tokens, another might not transcribe anything at all.

The algorithm expects the user to provide two aligned lists of strings, one with the models hypothesis and another one with the references, in which words of interest, in our case the embedded language, are tagged. In the example refernce \enquote{das mit den $<$tag bots$>$ glaub ich nicht} (I dont believe that thing with the bots), the word \enquote{bots} would be considered the point of interest.
Due to the nature of language pairs with different writing scripts such as Arabic/Mandarin-English code-switching, this is done automatically in the metrics.
For other commonly available code-switching test data such as DECM \cite{ugan2024decm} the language information is usually provided as well, and words can easily be tagged appropriately.

\section{Evaluating CSW with PIER}
\label{subsec:Evaluation-PIER}
In most cases, English is treated as the embedded language, and the matrix language is one of Spanish, Arabic, German, or Mandarin.
For that reason, we determine English words to be our points of interest and want to focus evaluation of model predictions to these words, asking: \emph{is the model able to do code-switching or does it get confused?}
For this evaluation, we excluded the monolingual utterances from the code-switching test set to avoid any contamination in the evaluation.
The left hand side of Table~\ref{tab:eval-comparison-per} shows the pre-trained models base PIER performance on our test data.
The first thing we can observe is that the PIER is notably higher than the general WER, indicating that these code-switched locations are naturally more challenging than the main languages in the test sets.


The right side of Table~\ref{tab:eval-comparison-per} shows the PIER results after fine-tuning and improving on our test data (in analogy to Table~\ref{tab:eval-comparison-wer}, right).
However, we can clearly see for several examples that while the WER improved after fine-tuning, the PIER increases.
While the W-large (agnostic) model showed a relative improvement of 5.06\% WER on the Fisher data, the PIER performance shows a relative worsening of 39.47\%. 
For a better visualisation of the phenomenon the blue lines in Figure~\ref{fig:comparison-wer-per} a) show an increase in the gap between WER and PIER after fine-tuning.
Another point where this evaluation reveals the truth about code-switching performance is on the SEAME.sge evaluation of the same model (orange lines).
\begin{table*}[t!]
\caption{Comparison of pre-trained and fine-tuned model PIER \(\downarrow\) evaluations on Fisher (ES-EN), Arzen (AR-EN), and SEAME (ZH-EN).}
\centering
\scriptsize
\setlength\tabcolsep{5px}
\renewcommand{\arraystretch}{1}
\begin{tabular}{|l|c|c|c|c|c||c|c|c|c|c|}
\hline
\multicolumn{1}{|c|}{\textbf{Model}} & \multicolumn{1}{c|}{\textbf{Decoding}} & \multicolumn{4}{c||}{\textbf{Pre-trained Model (PIER \(\downarrow\))}} & \multicolumn{4}{c|}{\textbf{Fine-tuned Model (PIER \(\downarrow\))}} \\
\hline
 &  & \textbf{Fisher} & \textbf{Arzen} & \textbf{SEAME.man} & \textbf{SEAME.sge} & \textbf{Fisher} & \textbf{Arzen} & \textbf{SEAME.man} & \textbf{SEAME.sge} \\
\hline
W-large & agnostic & 36.41 & 58.99  & 58.73 & 64.51 & 50.78 & 57.18  & 72.49 & 68.49 \\
              & EN  & 33.31 & 38.77 & 39.90 & 36.10 & 36.95 & 38.72 & 71.26  & 67.36 \\
              & X  & 59.92 & 70.78  & 60.23 & 76.26 & 51.16 & 58.63 & 72.49  & 68.49\\
\hline
W-small & agnostic & 48.14 & 75.91 & 74.95 & 83.08 & 63.52 & 78.10 & 88.82 & 97.60\\
              & EN  & 42.56 & 45.16 & 45.02 & 44.55 & 51.63 & 44.91  & 44.55 & 42.34  \\
              & X  & 62.55 & 80.66  & 72.88 & 77.13 &  63.87 & 79.57  & 90.24 & 100.03 \\
\hline
MMS     & EN & 74.90 & 79.80 & 86.48 & 86.08 & -- & -- & -- & -- \\
        & X & 75.29 & 100.52 & 109.78 & 105.70 & 71.73 & 89.68 & 108.61  & 103.97\\
\hline
\end{tabular}
\label{tab:eval-comparison-per}
\end{table*}
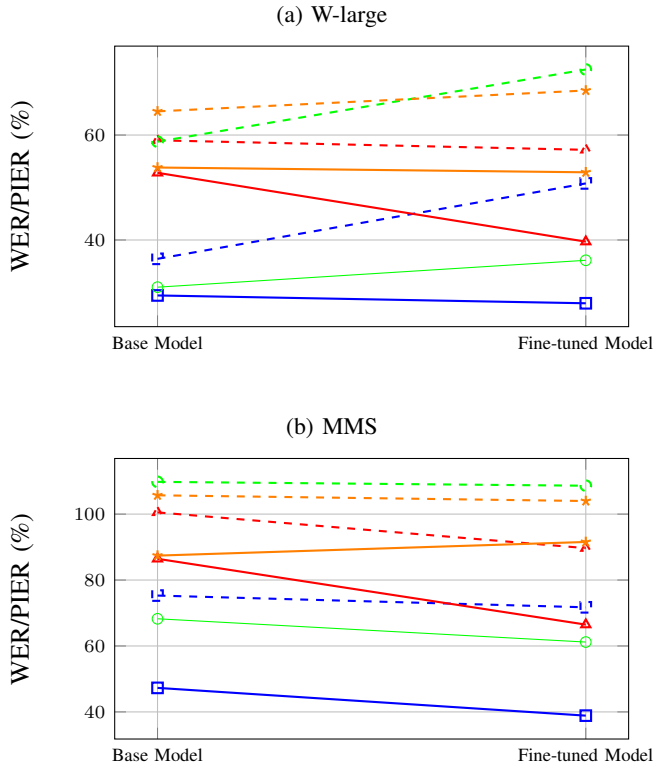
\begin{figure}[t!]
    \centering
    \begin{minipage}{\linewidth}
        \centering
        \begin{tikzpicture}
    \begin{axis}[
        axis lines=none, 
        xlabel=\empty, 
        ylabel=\empty, 
        legend style={
            at={(0.5,-0.1)}, 
            anchor=south,
            legend columns=4,
            font=\scriptsize, 
            cells={anchor=west}, 
            draw=none,
            /tikz/legend image code/.code={%
                \draw [line width=0.8pt] (0,0.5ex) -- (0.5ex,0.5ex); 
            },
        },
        width=0.9\linewidth, 
        height=0.2\linewidth, 
        tick style=none, 
        xtick=\empty, 
        ytick=\empty, 
        xmin=0,xmax=1,ymin=0,ymax=1,
        clip=false, 
    ]
        \addlegendimage{only marks, color=blue, mark=square, line width=0.8pt};
        \addlegendimage{only marks, color=red, mark=triangle, line width=0.8pt};
        \addlegendimage{only marks, color=green, mark=o, line width=0.8pt};
        \addlegendimage{only marks, color=orange, mark=star, line width=0.8pt};
        
        \addlegendentry{Fisher}
        \addlegendentry{ARZEN}
        \addlegendentry{SEAME.MAN}
        \addlegendentry{SEAME.SGE}
    \end{axis}
\end{tikzpicture}
\begin{tikzpicture}
    \begin{axis}[
        axis lines=none, 
        xlabel=\empty, 
        ylabel=\empty, 
        legend style={
            at={(0.5,-0.1)}, 
            anchor=south,
            legend columns=2,
            draw=none,
            font=\scriptsize, 
            cells={anchor=center}, 
            /tikz/legend image code/.code={%
                \draw [line width=0.8pt] (0,0.5ex) -- (0.5ex,0.5ex); 
            },
        },
        width=\linewidth, 
        height=0.2\linewidth, 
        tick style=none, 
        xtick=\empty, 
        ytick=\empty, 
        xmin=0,xmax=1,ymin=0,ymax=1,
        clip=false, 
    ]
        \addlegendimage{color=black, mark=none, line width=0.8pt};
        \addlegendimage{color=black, mark=none, dashed, line width=0.8pt};
        
        \addlegendentry{WER}
        \addlegendentry{PIER}
    \end{axis}
\end{tikzpicture}
    \end{minipage}\hfill
    \vspace{-15pt}
    \begin{minipage}{\linewidth}
        \centering
        \subcaption{W-large}
        \begin{tikzpicture}
    \begin{axis}[
    ylabel={WER/PIER (\%)},
    legend style={
        at={(1,1.15)},
        anchor=south east,
        legend columns=2,
        font=\footnotesize, 
        cells={anchor=west}, 
        /tikz/legend image code/.code={%
            \draw [line width=0.8pt] (0,0.5ex) -- (0.5ex,0.5ex); 
        },
    },
    xtick={1,2},
    xticklabels={Base Model, Fine-tuned Model},
    tick label style={font=\scriptsize}, 
    grid=major,
    width=0.95\linewidth, 
    height=0.6\linewidth, 
    ]
        \addplot[
            color=blue,
            mark=square,
            line width=0.8pt,
            ]
            coordinates {(1,29.43) (2,27.94)};
        \addplot[
            color=blue,
            dashed,
            mark=square,
            line width=0.8pt,
            ]
            coordinates {(1,36.41) (2,50.78)};
        
        \addplot[
            color=red,
            mark=triangle,
            line width=0.8pt,
            ]
            coordinates {(1,52.80) (2,39.68)};
        \addplot[
            color=red,
            dashed,
            mark=triangle,
            line width=0.8pt,
            ]
            coordinates {(1,58.99) (2,57.18)};
        
        \addplot[
            color=green,
            mark=o,
            ]
            coordinates {(1,31.01) (2,36.13)};
        \addplot[
            color=green,
            dashed,
            mark=o,
            line width=0.8pt,
            ]
            coordinates {(1,58.73) (2,72.49)};
        
        \addplot[
            color=orange,
            mark=star,
            line width=0.8pt,
            ]
            coordinates {(1,53.81) (2,52.90)};
        \addplot[
            color=orange,
            dashed,
            mark=star,
            line width=0.8pt,
            ]
            coordinates {(1,64.51) (2,68.49)};
        

    \end{axis}
\end{tikzpicture}
    \end{minipage}\hfill
   \vspace{-8pt}
   \begin{minipage}{\linewidth}
       \centering
       \subcaption{MMS}
       \begin{tikzpicture}
    \begin{axis}[
    ylabel={WER/PIER (\%)},
    xtick={1,2},
    xticklabels={Base Model, Fine-tuned Model},
    tick label style={font=\scriptsize}, 
    grid=major,
    width=0.95\linewidth, 
    height=0.6\linewidth, 
    ]
        \addplot[
            color=blue,
            mark=square,
            line width=0.8pt,
            ]
            coordinates {(1,47.29) (2,38.87)};
        \addplot[
            color=blue,
            dashed,
            mark=square,
            line width=0.8pt,
            ]
            coordinates {(1,75.29) (2,71.73)};
        
        \addplot[
            color=red,
            mark=triangle,
            line width=0.8pt,
            ]
            coordinates {(1,86.44) (2,66.48)};
        \addplot[
            color=red,
            dashed,
            mark=triangle,
            line width=0.8pt,
            ]
            coordinates {(1,100.52) (2,89.68)};
        
        \addplot[
            color=green,
            mark=o,
            ]
            coordinates {(1,68.25) (2,61.19)};
        \addplot[
            color=green,
            dashed,
            mark=o,
            line width=0.8pt,
            ]
            coordinates {(1,109.78) (2,108.61)};
        
        \addplot[
            color=orange,
            mark=star,
            line width=0.8pt,
            ]
            coordinates {(1,87.41) (2,91.55)};
        \addplot[
            color=orange,
            dashed,
            mark=star,
            line width=0.8pt,
            ]
            coordinates {(1,105.70) (2,103.97)};
        

    \end{axis}
\end{tikzpicture}
   \end{minipage}
    \vspace{-25pt}
    \caption{Comparison of WER and PIER behaviour after finetuning base models. (a) W-large, (b) MMS. Values are taken from the agnostic decoding for (a).}
    \label{fig:comparison-wer-per}

\vspace{-15pt}
\end{figure}
While the relative WER improvement was 1.69\% the performance on PIER decreased by 6.17\%.
These numbers suggest that evaluating PIER instead of WER provides more valuable insights on how the model actually behaves on the code-switched parts of the utterance.

\smallheading{Fine-grained evaluation}
For languages like German \& Arabic, intra-word code-switching is a common occurrence.
This form of code-switching is even more difficult than inter-word switching, because new words are created by using parts of each language.
Using PIER enables us to attribute improvements to the different kinds of code-switching by defining only intra- or inter-word code-switched cases as points of interest.
This way, more focused and clearer contributions can be made.

For DECM (German-English), we extended the data by manually annotating whether a word is English, German, a name, or an intra-word case of CSW. 
Arzen (Arabic-English) also provides necessary annotations to determine any intra-word switching.
Table~\ref{tab:PIER-inter-intra-word} shows our pre-trained models PIER performance on inter-word and intra-word level code-switching.
The ASR models significantly struggle for both cases, but it is clear that intra-word level code-switching is significantly more challenging.
\begin{table}[t!]
\caption{PIER \(\downarrow\) of different pre-trained models across DECM and Arzen datasets. Evaluation for inter-word (Inter.W) and intra-word (Intra.W) code-switching.}
\centering
\scriptsize
\begin{tabular}{|l|c|c|c|c|c|}
\hline
{Model} & Decoding & \multicolumn{2}{c|}{DECM (PIER \(\downarrow\))} & \multicolumn{2}{c|}{Arzen (PIER \(\downarrow\))} \\ \cline{3-6} 
                    &   & Inter.W & Intra.W & Inter.W & Intra.W \\ \hline
W-large    &agnostic            &   30.42    &   45.23    & 49.50,                                              
                                                        
                 & 105.71 \\ \hline
W-small   &agnostic & 39.56 & 73.32 & 69.12 &   107.45 \\ \hline
MMS     &X   &  64.22  &  83.26  & 100.46 &  100.79   \\ \hline
\end{tabular}
\label{tab:PIER-inter-intra-word}
\vspace{-8pt}
\end{table}
\section{Conclusion}
In this work, we introduced PIER, a metric for evaluating and analyzing code-switching ASR in a standardized and more expressive way.
More importantly, our proposed metric also enables to quantify the amount of errors for different types of code-switching, such as inter-\& intra-word level.
To the best of our knowledge, this is the first attempt to separate code-switching specific errors within a test set, and in several datasets we can also identify and quantify the errors for intra-word and inter-word types of CSW.
By doing so, we can show that, typical practices that improve WER in general, such as fine-tuning on in-domain data that sounds like an obvious way to improve, do not get closer to solving this problem even with models at the caliber of Whisper.
We hope that our work enables a more sophisticated evaluation of models when it comes to specific tasks wherein points of interest are overshadowed by some other majority, such as the matrix language in the case of code-switching speech.
This will also enable researchers to tackle and analyze different difficulties arising in code-switching speech in a more pinpointed way.

\section{Acknowledgment}
This work was funded by the German Federal Ministry of Education and Research (BMBF) under grant 01EF1803B (RELATER), the pilot program Core-Informatics of the Helmholtz Association (HGF) and the EU Horizon program (grant 101135798, Meetween), the HoreKa supercomputer funded by the Ministry of Science, Research and the Arts Baden-Württemberg and BMBF, also partly DeltaGPU at NCSA through allocation CIS240493 from the Advanced Cyberinfrastructure Coordination Ecosystem: Services \& Support (ACCESS) program, supported by U.S. National Science Foundation grants \#2138259, \#2138286, \#2138307, \#2137603, and \#2138296\cite{access}. 


\end{document}